\documentclass[letterpaper, 10 pt, conference]{ieeeconf}  % Comment this line out if you need a4paper

\IEEEoverridecommandlockouts                              % This command is only needed if 
\usepackage{graphics} % for pdf, bitmapped graphics files
\usepackage{epsfig} % for postscript graphics files
\usepackage{mathptmx} % assumes new font selection scheme installed
\usepackage{times} % assumes new font selection scheme installed
\usepackage{amsmath} % assumes amsmath package installed
\usepackage{amssymb}  % assumes amsmath package installed
\usepackage{float}
\usepackage{subfigure}
\usepackage{booktabs}

\title{\LARGE \bf
Separating Drone Point Clouds From Complex Backgrounds by Cluster Filter -- Technical Report for CVPR 2024 UG2 Challenge
}

\author{Hanfang Liang$^{1}$, Jinming Hu$^{2}$, Xiaohuan Ling$^{3}$, Bing Wang$^{4}$% <-this % stops a space
\thanks{*This work was not supported by any organization}% <-this % stops a space
\thanks{$^{1}$Hanfang Liang is with the School of Jianghan University , Wuhan, China.
        {\tt\small hanfangliang@stu.jhun.edu.cn}}%
\thanks{$^{2}$Jinming Hu is with the School of Jianghan University , Wuhan, China.
        {\tt\small 18571861306@stu.jhun.edu.cn}}%
\thanks{$^{3}$Xiaohuan Ling is with the School of Jianghan University , Wuhan, China.
        {\tt\small 1586731584@qq.com}}%
\thanks{$^{4}$Bing Wang is the corresponding author with the School of Jianghan University , Wuhan, China.
        {\tt\small  wangbing@jhun.edu.cn}}%
}

\begin{document}

\maketitle
\thispagestyle{empty}
\pagestyle{empty}

% Force show page number
\thispagestyle{plain}
\pagestyle{plain}
%%%%%%%%%%%%%%%%%%%%%%%%%%%%%%%%%%%%%%%%%%%%%%%%%%%%%%%%%%%%%%%%%%%%%%%%%%%%%%%%
\begin{abstract}
%Recently, target detection of small UAVs has received more and more attention due to the widespread application of UAVs. However, for situations where the target is small, far away, the point cloud is sparse, or the background is complex, although deep learning methods perform well in many target detection tasks, in the dominant task of unsupervised UAV detection, deep learning methods still have facing difficulty. This paper proposes a UAV detection method using an unsupervised method. It uses spatial-temporal sequence processing to fuse multiple lidar data to effectively track and determine the position of UAVs, so as to detect and track UAVs in challenging environments. machine. Our method performs front and rear background segmentation of point clouds through a global-local sequence clusterer, and parses point cloud data from both the spatital-temporal density and spatial-temporal voxels of the point cloud. Furthermore, a scoring mechanism for point cloud moving targets is proposed, using time series detection to improve accuracy and efficiency. We use the MMAUD dataset, and the method achieved 4th place in the CVPR 2024 UG2+ Challenge, confirming the effectiveness of our method in practical applications.

The increasing deployment of small drones as tools of conflict and disruption has amplified their threat, highlighting the urgent need for effective anti-drone measures. However, the compact size of most drones presents a significant challenge, as traditional supervised point cloud or image-based object detection methods often fail to identify such small objects effectively. This paper proposes a simple UAV detection method using an unsupervised pipeline. It uses spatial-temporal sequence processing to fuse multiple lidar datasets effectively, tracking and determining the position of UAVs, so as to detect and track UAVs in challenging environments. Our method performs front and rear background segmentation of point clouds through a global-local sequence clusterer and parses point cloud data from both the spatial-temporal density and spatial-temporal voxels of the point cloud. Furthermore, a scoring mechanism for point cloud moving targets is proposed, using time series detection to improve accuracy and efficiency. We used the MMAUD dataset, and our method achieved 4th place in the CVPR 2024 UG2+ Challenge, confirming the effectiveness of our method in practical applications.
%Index Terms—Global-local, MAV detection, Cluster, point cloud.

\end{abstract}

%%%%%%%%%%%%%%%%%%%%%%%%%%%%%%%%%%%%%%%%%%%%%%%%%%%%%%%%%%%%%%%%%%%%%%%%%%%%%%%%
\section{INTRODUCTION}

Drones have received huge attention in various real-world applications, such as surveillance, military applications, surveying and mapping monitoring, etc. Regarding drone monitoring, currently there are mainly vision-based detection solutions. 
However, when the UAV is far away from the target, it is difficult to detect the position of the UAV solely by relying on images, and it is also difficult to obtain the spatial location information of the UAV solely by relying on images.

\begin{figure}[h]
\centering
\subfigure[Sparse MAV Point Cloud]{
\includegraphics[width=6cm, height=6cm]{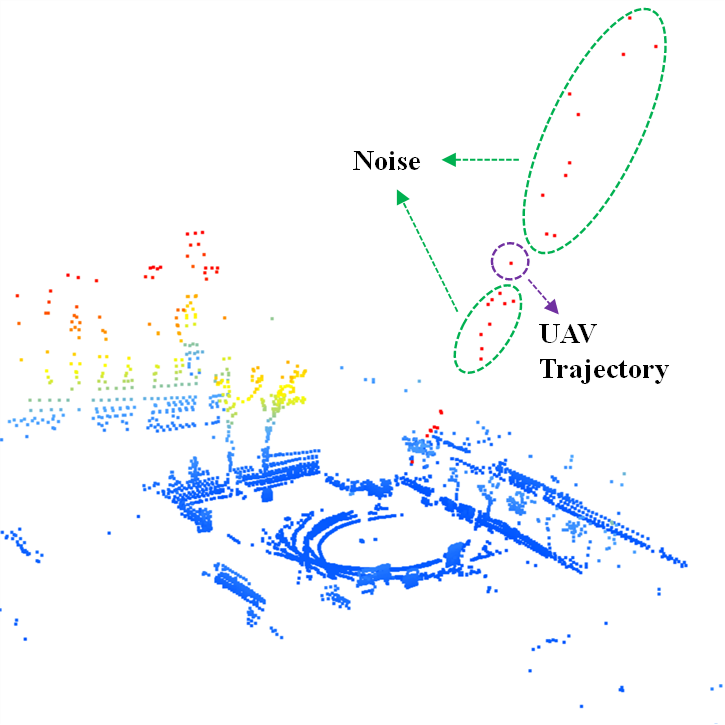}}
\subfigure[Small MAV (7×7 pixels)]{
\includegraphics[width=6cm, height=5cm]{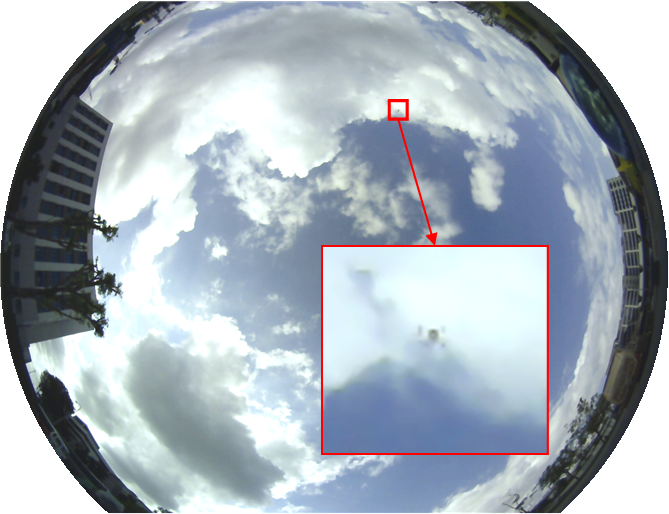}}
\caption{Challenging examples of image and point cloud detection. In the point cloud, the scanning points of the drone are very sparse and not continuous in the time dimension. In many time frames, the drone is too small to be detected. In the picture, the drone is very tiny, only a dozen pixels.}
\vspace{-15pt}
\label{1}
\end{figure}

Compared with the two-dimensional image detection of UAVs, the three-dimensional spatial position detection and estimation of UAVs is more challenging. It is not only necessary to detect and track the target UAV, but also to estimate the spatial position of the UAV. The main challenges are: first, when the drone is flying in a high-altitude orbit, the image features of the drone are weak and only occupy a few pixels. At the same time, it is easily affected by the environment such as lighting and is difficult to identify. Secondly, the point cloud of the drone scanned by lidar is sparse, and the drone cannot be scanned in every frame, resulting in the point cloud characteristics of the drone being very unstable. Third, point cloud data is provided by multiple lidars, and some of the lidar data contain a lot of noise, making it difficult to correctly identify and track drones.

Our goal is to build an unsupervised UAV point cloud detection method, segment the point cloud of the UAV trajectory from the unknown point cloud space, retain only the point cloud containing the UAV trajectory, and restore it combined with the timestamp information. The original trajectory of the drone.

In this article, we mainly classify the point cloud point sets through the clustering idea. In particular, we divide the clusterer into two parts. The global-local clusterer classifies the overall point cloud, and initially divides and excludes buildings. Large objects such as objects and trees; then based on the clustering results, the two attributes of spatiotemporal density and spatiotemporal voxels are further separated, and the moving targets in the spatiotemporal sequence are further processed to select the point cloud corresponding to the UAV trajectory. Finally, a filter spline fitting operation is performed on the point cloud, and the spatial position of the UAV is restored based on the timestamp interpolation.

The main contributions of our work are as follows:

\begin{itemize}

\item We provide a simple and fast unsupervised detection method for detecting drone trajectories and positions from point cloud data. Our method only uses point cloud sources to detect drones and only uses lidar data in the MMAUD dataset. It does not rely on complex deep learning algorithms and can be quickly deployed in edge devices.
\item We propose a spatio-temporal voxel and spatio-temporal density analysis method for point cloud moving targets, and a scoring mechanism to evaluate the confidence of the point cloud to isolate the correct trajectory point set.
\item Our method achieved 4th place in the CVPR 2024 UG+ Challenge, confirming the effectiveness and reliability of our method for tracking UAVs and determining the spatial location of UAVs.
\item We compared and selected some deep learning algorithms and conducted ablation experiments to formalize the speed and usefulness of our method for UAV detection.

\end{itemize}

%%%%%%%%%%%%%%%%%%%%%%%%%%%%%%%%%%%%%%%%%%%%%%%%%%%%%%%%%%%%%%%%%%%%%%%%%%%%%%%%
\section{Related Works}

\begin{figure*}[t]
\centering
\includegraphics[width=\textwidth]{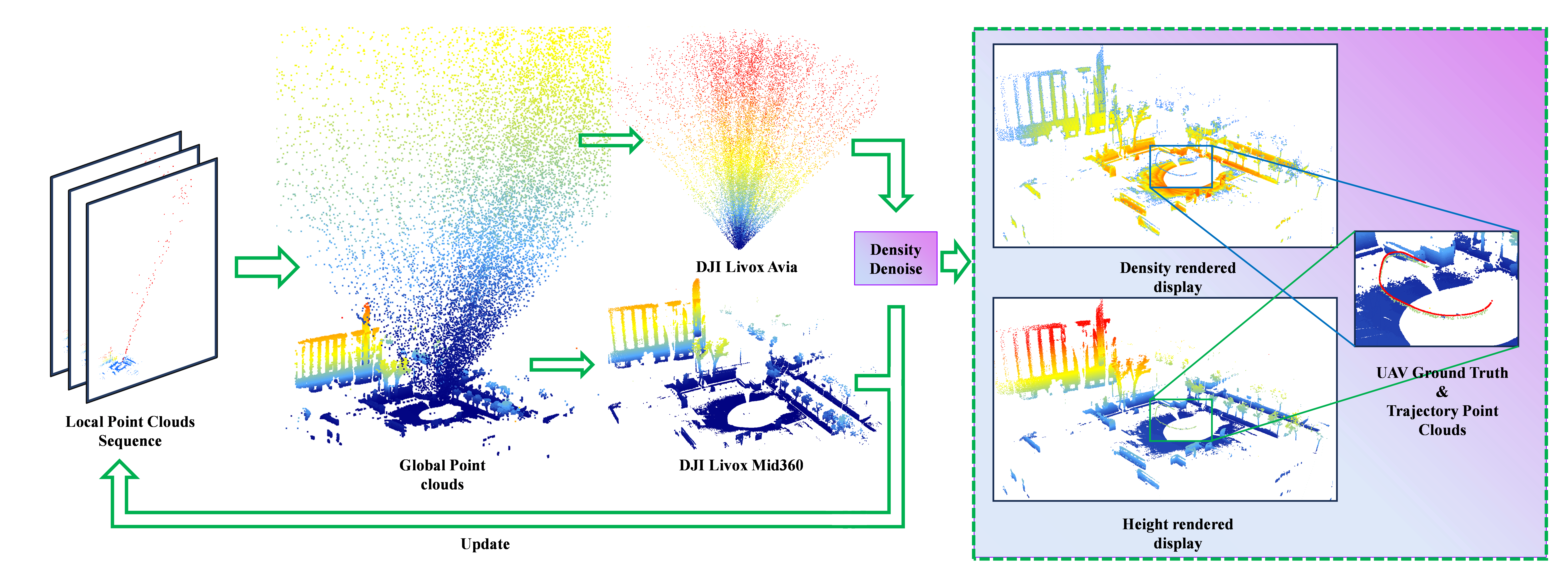}
\caption{We first superimpose all point clouds in the sequence to obtain the global point cloud, then separate the data from different lidars, and only perform denoising processing on the data from DJI Livox Avia. The picture in the box on the right is radar data that has been processed by noise reduction, and is rendered and distinguished according to point cloud density and point cloud height. The greater the density of the point cloud and the higher the altitude, the more red the color becomes; the red trajectory in the picture on the right is the real trajectory of the drone. It can be seen that the noise is well processed and the trajectory point cloud of the drone is preserved.}
\label{fig:twocolumn}
\end{figure*}

This section reviews the literature in the field of UAV
detection and tracking. Due to the limited research on tracking
small objects such as UAVs based on LiDAR, we focus on:
(A) vision-based and (B) LiDAR-based UAV tracking; (C)

\subsection{Vision-based UAV tracking}
There are currently many vision-based UAV detection methods and have received attention in many applied tasks. With the rapid development of deep learning, there have been many studies on UAV detection methods based on deep learning. Y. Zheng et al. evaluated eight state-of-the-art deep learning algorithms for MAV detection on the Det-Fly dataset\cite{zheng2021air}. B. K. S. Isaac-Medina et al. evaluated four state-of-the-art deep learning algorithms on three representative MAV datasets (MAV-VID, Drone-vs-Bird, Anti-UAV)\cite{isaac2021unmanned}. In order to further improve the accuracy of target detection, H. Liu et al. implemented a special data augmentation method and pruned the convolution channels and skip layers of YOLOv4 for small UAV detection\cite{liu2021real}. C. Rui et al. proposed a novel comprehensive method that combines transfer learning and adaptive fusion based on simulated data to improve small target detection performance\cite{rui2021comprehensive}.

Motion-assisted micro air vehicle (MAV) detection methods aim to detect MAVs by combining motion features and appearance features. Existing motion-assisted MAV detection methods can be divided into two categories: fixed cameras and mobile cameras. U. Seidaliyeva et al. used a fixed camera to monitor the sky and then used background subtraction and CNN-based object classification for MAV detection\cite{seidaliyeva2020real}. J. Xie et al. used a method to fuse the spatiotemporal characteristics of the target for the detection of long-range flying drones\cite{xie2021small}-\cite{xie2020adaptive}.
Y. Zheng et al. used appearance features to exclude non-MAV moving targets, and then used motion-based classification algorithms to distinguish MAVs from other distractors\cite{zheng2022detection}.

MAV detection from moving cameras is more challenging than from fixed cameras because the motion of the background is coupled with the motion of the target. J. Li et al. proposed a UAV-to-UAV video dataset and a general architecture for small MAV detection from cameras mounted on mobile MAV platforms. The authors detect moving MAVs by subtracting adjacent frames and then use a hybrid classifier to identify MAVs\cite{li2021fast}.

M. W. Ashraf et al. proposed a two-stage segmentation method. In the first stage, the authors utilize 2D convolutional networks and channel-wise pixel-level attention to extract contextual information from overlapping patches. Then, a 3D convolutional network and channel-pixel-level attention are used to learn spatiotemporal cues and discover first-stage omission detection\cite{ashraf2021dogfight}.
The method in [7] proposes a UAV detector based on feature super-resolution, which is based on motion information extraction of dense optical flow\cite{wang2023low}.
However, this is still challenging for air-to-air MAV detection in complex environments.

\subsection{LiDAR-based UAV tracking}

While lidar systems are commonly used to detect and track objects, unmanned aerial vehicles (UAVs) present unique challenges in detecting and tracking UAVs due to their small size, shape, diverse materials, high speeds, and unpredictable motion challenges {\cite{lei2024audio}}.

Li proposed a new method for tracking drones using lidar point clouds. They consider the speed and distance of the drone to adjust the lidar frame integration time, parameters that have an impact when dealing with the density and size of the point cloud\cite{qingqing2021adaptive}.

Sedat Dogru et al. suggest that detection can be accomplished using fewer lidar beams as long as a probabilistic analysis of detection is performed and appropriate settings are ensured. When tracking a small number of hit points continuously, the limitations of 3D lidar technology can be overcome by moving the sensor to increase the field of view and improve coverage\cite{dogru2022drone}. Razlaw, J. et al. proposed a method that combines segmentation methods and simple object models while utilizing temporal information to overcome the limitations of 3D lidar technology and improve UAV detection and tracking capabilities\cite{razlaw2019detection}. Wang, H et al. used Euclidean distance clustering and particle filtering algorithms to complete UAV detection and tracking\cite{wang2021study}. Sier et al. proposed the concept of lidar as a camera to track drones without any prior knowledge about the data content through the fusion of images and point cloud data generated by a single lidar sensor. Using a custom YOLOv5 model trained on panoramic images, they were able to integrate computer vision capabilities directly onto the lidar itself\cite{sier2023uav}.

Although deep learning methods have made great progress relative to traditional methods, these methods are either too time-consuming or only effective when the target is large enough or the background is very simple, and for different scenarios and different drone types , data-driven deep learning methods require large UAV data sets.
Therefore, there are still many challenges in the deep learning MAV detection method based on image appearance, such as difficulty in detection in complex background environments and difficulty in detection of small objects. At the same time, for image-based detection, it is difficult to estimate the three-dimensional position of the drone, making it difficult to meet the needs of many practical applications.

The use of lidar technology offers multiple ways to improve drone detection and tracking and explore new technologies to overcome the unique challenges posed by these small, fast-moving and unpredictable objects. But at the same time, the lidar point cloud detection method also needs to face problems such as point cloud noise, the sparsity of the lidar point cloud, and the discontinuity of small targets in the point cloud sequence{\cite{liang2024unsupervised}}. And compared to detecting the position of the drone in the image, this article focuses on using unsupervised methods to detect the pose and trajectory of the drone, as well as detecting and predicting the spatio-temporal coordinates of the drone{\cite{xiao2024tame}}.

%%%%%%%%%%%%%%%%%%%%%%%%%%%%%%%%%%%%%%%%%%%%%%%%%%%%%%%%%%%%%%%%%%%%%%%%%%%%%%%%
\section{Methods}

\begin{figure*}[t]
\centering
\includegraphics[width=\textwidth]{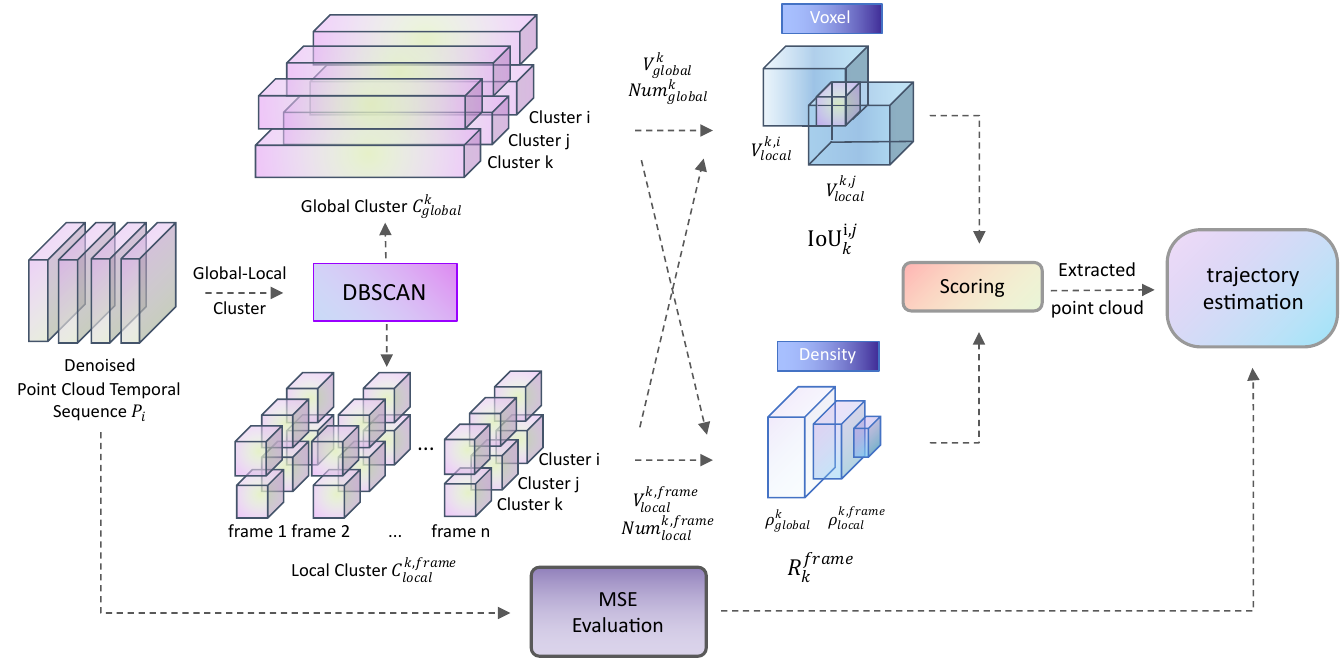}
\caption{Our proposed algorithm architecture. Given a continuous point cloud input sequence, we first classify it into global clustering and local clustering to obtain different categories using DBSCAN. Then the number of point clouds and voxel spatial information are calculated for global and local categories respectively. And cross-compare and calculate the spatial coincidence degree and temporal density changes of different clusters. The final scoring mechanism calculates the spatial coincidence degree and relative density score of the categories to exclude point clouds other than the UAV, restores the trajectory of the UAV through spline fitting interpolation, and uses MSE to calculate the error with the true value.}
\label{fig:twocolumn}
\end{figure*}

This section presents the details of our proposed method. To effectively detect MAVs under challenging conditions, we propose a clustering-based point cloud unsupervised spatial-temporal sequence UAV trajectory detection method. Consists of three parts. In Sect. 3.1, We first denoise the point clouds data from different LiDARs; In Sect. 3.2, we design a global clusterer and a local clusterer; In Sect. 3.3, we will introduce a scoring mechanism to evaluate the confidence of each clustering result and filter the clustering categories for separation. UAV trajectory; In Sect. 3.4, we combine the processed UAV point cloud with time frame regression to fit the UAV spatial position.

\subsection{Point clouds denoise}
The detected point cloud data comes from two lidar sensors, DJI Livox Avia and DJI Livox Mid360. By combining the two radars, a lidar scanning range close to full coverage from the ground to the sky is obtained. However, lidar data is a sparse signal, and the DJI Livox Avia radar will be accompanied by a lot of noise, which can range up to 400 meters, and the small targets of the drone itself are very similar to the noise. If the data with a lot of noise is used directly, it will leading to disastrous consequences.
Therefore, we first need to denoise the lidar data from DJI Livox Avia.

We first superimpose the continuous lidar sequences and superimpose all the sequences. We can find that the noise density from the sensor belongs to the sparsest category. Based on this, we exclude noise points based on the density, while ensuring high accuracy. as shown in picture 2,.
The excluded noise points are then updated to the lidar sequence to facilitate the use of subsequent local and global clustering methods, which will be introduced in the next section.

\subsection{Global-local point set clusterings}
Lidar data is a sparse signal accompanied by a lot of noise. For the same stationary object, the points scanned at different times still have large spatial differences. However, as the time dimension increases, the point set density on the stationary object increases. will increase significantly. Inspired by some previous clustering ideas, the point set is initially divided into point sets of different densities through density and distance clustering.

In DBSCAN, the density associated with a point is obtained by counting the number of points within a specified radius area around the point. Points with density above a specified threshold are constructed as clusters. Among the existing clustering algorithms, we chose the DBSCAN algorithm because of its ability to discover clusters of arbitrary shapes, such as linear, concave, elliptical, etc. Moreover, compared with some clustering algorithms, it does not require the shape of the clusters to be determined in advance. DBSCAN's proven ability to handle large data.

A key issue in this method is how to correctly exclude point sets with different densities and retain the correct point set containing only UAV trajectories.
Considering that stationary object surfaces will accumulate more and more lidar point clouds as the time dimension increases. For moving objects of the same volume, the probability of being scanned by the lidar is the same in continuous time periods. Therefore, the point cloud density of a moving object should be relatively stable in continuous time frames.
In addition, for moving objects, the volume of the point cloud in the voxel space will change as the time dimension changes. Therefore, our method processes the spatio-temporal sequence from the two perspectives of point cloud density and voxel volume, and filters out the point cloud of the UAV trajectory.

Define the total number of frames in each sequence as n, The point set of each frame is \(P _ {i} , i\in \left ( 1,2,... ,n\right )\), The overall point set is \(C_{global}^{k}\).

\[P_{global}= \sum _{i= 1}^{n}P_{i},i\in \left ( 1,2,... ,n\right )\]
The number of frames the local point set contains in the time frame is frames, the starting frame of the local point set is set to frame, and \(frame\in \left ( 1,2,... ,n-frames\right )\).The local point set is \(P_{local}^{frame}\).

\[P_{local}^{frame}= \sum_{i=frame}^{frame+frames}P_{i} \]

We define the category k of the cluster in the global point set \(P_{global}\) as \(C_{global}^{k}\), the category k of the cluster in the local point set \(P_{local}^{frame}\) as \(C_{local}^{k,frame}\). In particular, the point set in \(C_{local}^{k,frame}\) is not a new cluster set obtained by DBSCAN re-clustering, but the corresponding points in \(C_{global}^{k}\) within a period of time in \(frame\in \left ( 1,2,... ,n-frames\right )\).

The Voxel space of the category k of the cluster \(C_{global}^{k}\) in global point set  \(C_{global}^{k}\) as \(V_{global}^{k}\), The Voxel space of the category k of the cluster \(C_{local}^{k,frame}\) in the local point set \(P_{local}^{frame}\) as \(V_{local}^{k,frame}\). The number of points of the category k of the cluster in local point set \(P_{local}^{frame}\) that changes over time is \(Num_{local}^{k,frame}\).

We first superimpose the point cloud on the global time frame to obtain \(P_{global}\), and use DBSCAN to perform clustering to obtain \(C_{global}^{k}\). We define and use closer and denser clustering parameters \(\epsilon \) and MinPts, and record the volume of each class at the same time. Prime information V. The global density of the point set is also recorded. Although a low threshold will bring some false targets, the most reliable targets are obtained by applying the confidence ranking and scoring mechanism, which will be introduced in detail in Section 3.2.

We calculate the density \(\rho_{global}^{k} \) of the point cloud in point set \(P_{global}\) according to \(C_{global}^{k}\).
\[\rho_{global}^{k} = \frac{Num_{global}^{k}}{V_{global}^{k}}\]

Considering that the target MAV will not cause drastic changes in spatial position in continuous time, a local point clusterer is used at this time.
for local point clusterer, First calculate the density \(\rho_{local}^{k,frame} \) of the point cloud in point set \(P_{local}^{frame}\) according to \(C_{local}^{k,frame}\). And Simultaneously calculate the spatial Intersection over Union (IoU) of the overlapping areas of voxels. Define the IoU of voxels in category k of the cluster \(C_{local}^{k,frame}\) between frame i and j as \(IoU_{k}^{i,j}\).
\[\rho_{local}^{k,frame} = \frac{Num_{local}^{k,frame}}{V_{local}^{k,frame}}\]

\[IoU_{k}^{i,j} = \frac{V_{local}^{k,i} \cap V_{local}^{k,j}}{V_{local}^{k,i} \cup V_{local}^{k,j}} \]
\[i,j =1,2,...,n-frames,i\neq j \]
And Calculate the ratio of local density to global density as Relative density \(R_{k}^{frame}\).
\[R_{k}^{frame}=\frac{\rho_{local}^{k,frame}}{\rho_{global}^{k}}\]

At this point, through the global-local clusterer, the relative density of each cluster point set \(R_{k}^{frame}\) and the IoU of voxels \(IoU_{k}^{i,j}\) can be obtained.

\subsection{Scoring Mechanism}

In this section, we will use the density and voxel coincidence obtained in the previous section. For a moving object, in the adjacent time dimension, as its spatial position changes, the voxel position in the space will also change accordingly. The local voxel coincidence should be smaller than the voxel coincidence of stationary objects. At the same time, as the time dimension increases, the density of point clouds accumulated on the surface of stationary objects will also increase. Therefore, the point cloud density of stationary objects will have a larger difference in global-local relative density; while the density of point clouds of moving objects will be It will be relatively consistent with the overall situation within a local time period.

Based on this inference, we designed a scoring mechanism to evaluate the confidence of point clouds in both density and voxel dimensions.And in order to make the value more stable and retain the changing trend of the value, we use the logarithmic function to map the voxel IoU to a more balanced scale.

We first define a voxel coincidence score for cluster k between local point set frame \(C_{local}^{k,frame}\) and \(C_{local}^{k,frame+1}\) as \(Score_{IoU}^{k}\).

\[Score_{IoU}^{k} = \sum_{k=1}^{n}log^{\frac{1}{IoU_{k}^{i,j}}}\]

Define the score of point set density matching between \(\rho_{global}^{k} \) and \(\rho_{local}^{k,frame} \) as \(Score_{dens}^{k}\).

\[Score_{dens}^{k} = \sum_{k=1}^{n} e^{R_{k}^{frame}}\]
The total score formula is recorded as \(Score^{k}\), where \(\lambda\) is  a hyperparameter set by us.

\[Score^{k} = Score_{dens}^{k} + \lambda \times Score_{IoU}^{k}\]

According to the calculation formula, the category with the highest score can be filtered out. Specifically, the target with the highest confidence is selected as the final target.

That is, the drone trajectory. In order to confirm the practicality of the confidence representation of the scoring mechanism, we randomly selected several groups of sequences to compare the separated trajectory point clouds with the true values, as shown in Figure (), which significantly removed cluttered point clouds such as background, while retaining the characteristics of the UAV Trajectory point cloud.

\subsection{Trajectory Prediction}

For the final trajectory based on time frame, we use spline fitting on the UAV point cloud, and then interpolate based on the time frame as the spatial position of the corresponding time frame.
For the time frame where the background is segmented, there may be multiple point clouds in the same frame. After data collection and sorting, even if there are multiple points corresponding to the same timestamp, they will be sorted into consecutive blocks in a list in chronological order. 

This means that for the same timestamp, all corresponding point cloud data are processed instead of just one point. Define the k-th point cloud frame after segmenting the background as \(P_{s}^{k}\). Sort the point clouds of each time frame according to the timestamp and merge them into a point set \(\mathbb{P}_{uav}=\left \{ P_{s}^{0}, P_{s}^{1},...,P_{s}^{k}\right \}\). Among them, the points in the point set \(\mathbb{P}_{uav}\) are selected as control points, and the three-dimensional spline S(u) can be expressed as:
\[S\left (  u\right )= \sum_{j=0}^{k}\sum_{i=0}^{n}P_{s}^{j}\left ( i \right )B_{i}\left ( u \right )\]
The basis function of the spline function is defined as:
\[B_{i}\left ( u \right )= \frac{1}{6}\left [ \left ( 1-u \right )^{3} ,3u^{3}-6u^{2}+4,-3u^{3}+3u^{2}+1,u^{3}\right ]\]
Finally, the three-dimensional curve is interpolated and fitted in the order of time frames to obtain the UAV spatial coordinates of the required time nodes.

\section{Experiment}

\subsection{Dataset}
To evaluate the performance of the proposed algorithm, we tested our proposed algorithm on MMAUD challenging dataset. MMAUD dataset is briefly introduced below.

The MMAUD dataset provides a multi-modal dataset that integrates
visual, LIDAR array, RADAR, and audio array sensors, and high-precision ground truth. The dataset contains over 1700 seconds of multimodal data divided into 50 different sequences. Each sequence contains sufficient visual, lidar, audio and radar data for identification purposes. Some example images are shown along with the visualized point cloud.

\subsection{Evaluation Metrics and Implementation Details}

In the CVPR 2024 UG2+ Challenge, we use MSE error to evaluate the accuracy of the algorithm. At the same time, we introduce the Sequence Detection Accuracy (SDA) indicator.
\[SDA=\frac{Detected Sequence Time}{All Sequence Time} \]
 Because for image sequences, there are situations where the target cannot be detected, and in this case, it is obviously more difficult to predict the spatial position coordinates of the drone. Therefore, when the target cannot be detected, predicting a completely irrelevant spatial position prediction coordinate will be meaningless, and the MSE of the method using camera modal data is very large.
Therefore, the SDA indicator also reflects the detection ability of different algorithms for small drones in complex backgrounds in this case.

The detection results of our method are shown in Figure 4. For the convenience of display, we superimpose the trajectories of the entire sequence. The green trajectory is the drone trajectory point cloud segmented from the background by global and local clustering methods; the red trajectory is the real spatial position of the drone; and the blue trajectory is the spatial position of the drone predicted by our method. Our method has achieved good results in eliminating noise and extracting the correct drone trajectory from the point cloud space.

\begin{figure}[htbp]
  \centering
  \subfigure[UAV Trajectory Point]{
  \includegraphics[width=6cm, height=6cm]{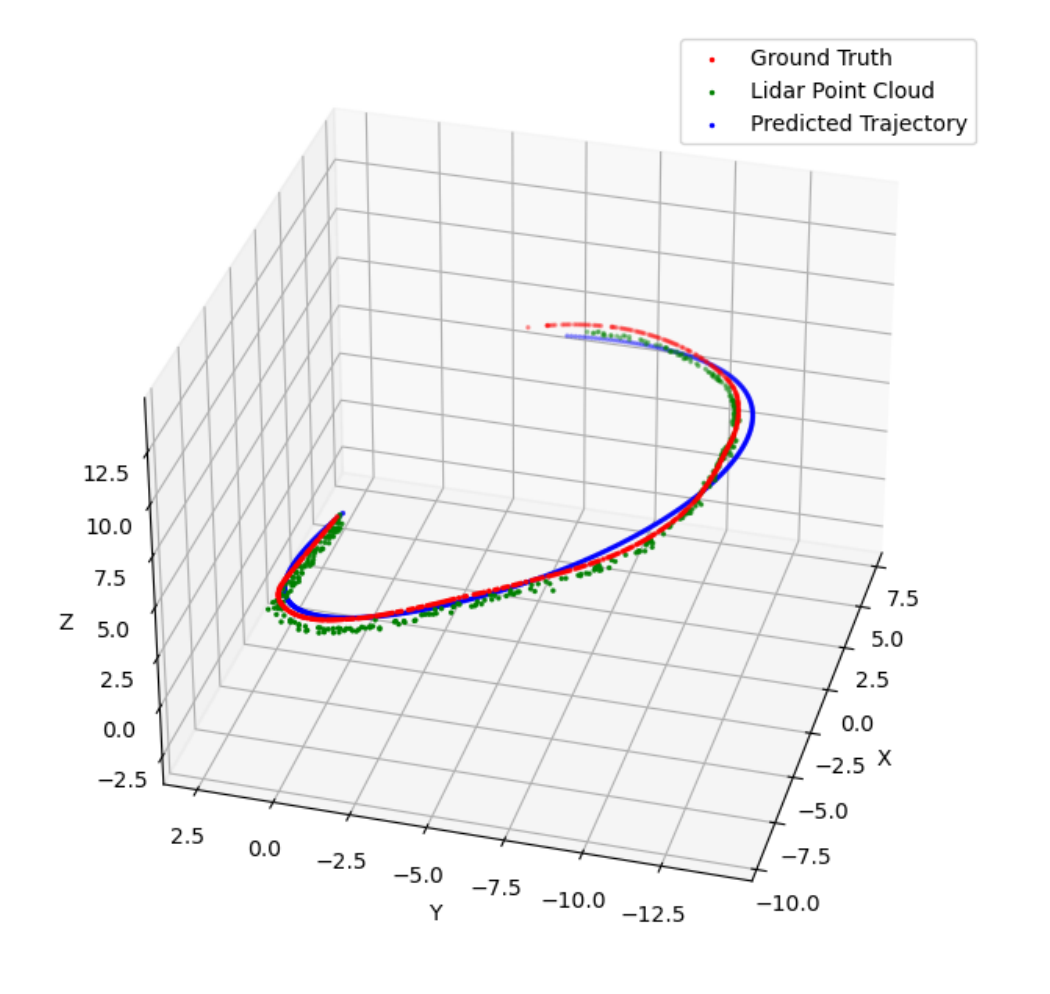}}
  
  \caption{The green point cloud in Figure b is the UAV point cloud separated from the background by our method, the red is the real trajectory of the UAV, and the blue is our predicted UAV trajectory.}
  \label{fig:example}
\end{figure}

\begin{table}
    \caption{Comparison of different methods}
\label{table_example}
    \label{num}
    \setlength{\tabcolsep}{2mm}{
    \begin{tabular}{ccccc}
    \hline
        \toprule
        methods & Modality & backbone & SDA & MSE \\
        \midrule
        \textbf{YOLOv5s}  & Image  & Resnet   & 1 & 1 \\
        \textbf{Cascade R-CNN }    & Image   & 20  & 1 & 1  \\
        \textbf{FPN R-CNN }    & Image   & 20  & 1 & 1  \\
        \textbf{Grid R-CNN }    & Image   & 20  & 1 & 1  \\

        \textbf{ViT}  & Image  & 20   & 1 & 1 \\
        \textbf{ }    & Image   & 20  & 1 & 1  \\
   
        \textbf{CenterNet}  & Point Cloud   & 20  & 1 & 1 \\
        \textbf{ }    & Point Cloud   & 20  & 1 & 1 \\
        
        \textbf{PointNet}  & Point Cloud   & 20  & 1 & 1 \\  
        \textbf{ }    & Point Cloud   & 20  & 1 & 1 \\

        \textbf{Ours}  & Point Cloud   & NA  & 99.15\(\%\) & 1 \\ 
        \bottomrule 
    \end{tabular}}
\end{table}

\section{CONCLUSIONS}

In this paper, we propose an unsupervised approach to MAV point cloud detector for ground-to-space detection of MAVs under challenging conditions. Our method employs spatial-temporal global local clustering of point cloud sequences for extracting effective UAV point cloud trajectories from sparse and noisy point clouds.

Our method won the 4th place in the CVPR2024 UG2+ Challenge, confirming the effectiveness of our method. Moreover, our method is interpretable. At the same time, we use the MMAUD dataset, evaluate several representative deep learning algorithms, and analyze the experimental results.

In the future, in order to detect different drones in the environment, the types of drones should be classified based on the distribution of drone point cloud trajectories and deep learning technology.

\addtolength{\textheight}{-12cm}   % This command serves to balance the column lengths

%%%%%%%%%%%%%%%%%%%%%%%%%%%%%%%%%%%%%%%%%%%%%%%%%%%%%%%%%%%%%%%%%%%%%%%%%%%%%%%%

%%%%%%%%%%%%%%%%%%%%%%%%%%%%%%%%%%%%%%%%%%%%%%%%%%%%%%%%%%%%%%%%%%%%%%%%%%%%%%%%

%%%%%%%%%%%%%%%%%%%%%%%%%%%%%%%%%%%%%%%%%%%%%%%%%%%%%%%%%%%%%%%%%%%%%%%%%%%%%%%%

%%%%%%%%%%%%%%%%%%%%%%%%%%%%%%%%%%%%%%%%%%%%%%%%%%%%%%%%%%%%%%%%%%%%%%%%%%%%%%%%

References are important to the reader; therefore, each citation must be complete and correct. If at all possible, references should be commonly available publications.

% \begin{thebibliography}{99}

% \end{thebibliography}

\bibliographystyle{IEEEtran}
%\bibliography{referernces}
\bibliography{mybib}

\end{document}